\documentclass[11pt]{article}

\usepackage[final]{acl}

\usepackage{times}
\usepackage{latexsym}

\usepackage[T1]{fontenc}

\usepackage[utf8]{inputenc}

\usepackage{microtype}

\usepackage{inconsolata}

\usepackage{graphicx}

\title{Lil: Less is Less When Applying Post-Training Sparse-Attention Algorithms in Long-Decode Stage}

\author{
  \textbf{Junhao Hu\textsuperscript{12*}},
  \textbf{Fangze Li\textsuperscript{3*}},
  \textbf{Mingtao Xu\textsuperscript{4}},
  \textbf{Feifan Meng\textsuperscript{3}},
  \textbf{Shiju Zhao\textsuperscript{3}},
  \textbf{Tiancheng Hu\textsuperscript{12}},\\
  \textbf{Ting Peng\textsuperscript{4}},
  \textbf{Anmin Liu\textsuperscript{12}},
  \textbf{Wenrui Huang\textsuperscript{3}},
  \textbf{Chenxu Liu\textsuperscript{12$^\dag$}},
  \textbf{Ziyue Hua\textsuperscript{12}},
  \textbf{Tao Xie\textsuperscript{215$^{\dag\ddagger}$}} \\
  \textsuperscript{1}SCS, Peking University, Beijing, China\\ 
  \textsuperscript{2}Key Lab of HCST (PKU), MOE, Beijing, China\\
  \textsuperscript{3}State Key Laboratory for Novel Software Technology, Nanjing University, China\\ 
  \textsuperscript{4}Tencent, Shenzhen, China \\
  \textsuperscript{5}Beijing Tongming Lake Information Technology Application Innovation Center, Beijing, China\\
}

\newcommand{\algo}{\textit{Guardian}}
\usepackage{booktabs}
\usepackage{algorithm}
\usepackage{algorithmic}
\usepackage{amsmath}
\usepackage{amssymb}
\usepackage[skip=0pt]{caption}
\usepackage{pifont}
\usepackage{multirow} 
\usepackage{titling}
\usepackage{balance}

\begin{document}
\maketitle
\renewcommand{\thefootnote}{\fnsymbol{footnote}}
\footnotetext[1]{Equal contribution. Author names appear in alphabetical order by last name.}
\footnotetext[2]{Corresponding authors.}
\footnotetext[3]{Tao Xie is with the Key Laboratory of High Confidence Software Technologies (Peking University), Ministry of Education; School of Computer Science, Peking University, Beijing, China;  Institute of Systems for Advanced Computing at Fudan University, Shanghai, China; Shanghai Institute of Systems for Open Computing, Shanghai, China.}
\renewcommand{\thefootnote}{\arabic{footnote}}

\begin{abstract}
Large language models (LLMs) demonstrate strong capabilities across a wide range of complex tasks and are increasingly deployed at scale, placing substantial demands on inference efficiency. Prior work typically decomposes inference into prefill and decode stages, with the decode stage dominating total latency, especially in reasoning-intensive tasks. To reduce time and memory complexity in the decode stage, a line of work introduces sparse-attention algorithms. In this paper, we show, both empirically and theoretically, that sparse attention can paradoxically increase end-to-end complexity: information loss often induces substantially longer sequences. We term this problem ``Less is Less'' (Lil). To mitigate the Lil problem, we propose an early-stopping algorithm that detects the threshold where information loss exceeds information gain during sparse decoding. Our early-stopping algorithm reduces token consumption by up to 90\% with a marginal accuracy degradation of less than 2\% across reasoning-intensive benchmarks.
\end{abstract}
\section{Introduction}
\label{sec-intro}

Large language models (LLMs)~\cite{openai2024o1, damai2024deepseek, mimo2025audio, mimo2025flash} exhibit strong capabilities across a wide range of complex tasks, such as software engineering~\cite{wang2023how, hu2023pcrml, liang2025diffedit, liu2025llmigrate},  writing~\cite{bai2024longbench}, and math solving~\cite{gsm8k, aime, math500}. Users interact with LLMs through natural-language-based prompts (sequences of tokens). This combination of capability and usability has driven rapid and widespread adoption of LLMs. As a result, LLMs must now be deployed at scale to handle an increasingly large and diverse set of requests, including long-input requests (e.g., document-question answering~\cite{bai2024longbench}), long-output requests (e.g., chain-of-thought reasoning~\cite{yao2023react} and long-form writing~\cite{bai2024longbench}, and code generation~\cite{wang2023how}), as well as requests requiring both~\cite{wu2025resum}. Longer inputs and outputs substantially increase inference latency and resource consumption, posing great challenges for large-scale deployment.

To address these inference challenges, prior work typically decomposes inference into two stages: prefill and decode. In the \textbf{prefill} stage, the model processes tokens given by users. It computes the Key (K) and Value (V) vectors for all tokens, stores these vectors in the KV cache, and generates the first output token to initiate the decode stage. In the \textbf{decode} stage, the model iteratively processes each newly generated token. It computes the KV vectors for the new token, appends these vectors to the KV cache, and generates the next token. This process repeats until a specified stopping criterion is met. This paper focuses on accelerating the decode stage, which dominates total inference time~\cite{hu2025raas, yao2023react}, especially in reasoning-intensive tasks.

To optimize the decode stage, a major line of work introduces sparse-attention algorithms\footnote{Although sparse attention also applies to prefill, this work focuses on the decode stage. We also emphasize post-training sparsity. Training-aware sparsity algorithms such as DeepSeek NSA (Native Sparse Architecture)~\cite{nsa2025yuan} fall outside our scope and are discussed in related work.}, aiming to reduce both time and memory complexity~\cite{zhang2023h2o, xiao2023sink, tang2024quest, hu2025raas, chen2024arkvale}. First, sparse attention reduces time complexity. Full attention requires each decode token to attend to all previous tokens. In contrast, sparse attention requires each decode token to attend to only the top-k most relevant tokens, substantially reducing computation while maintaining accuracy. Second, sparse attention may reduce memory complexity. Some algorithms reduce memory by discarding irrelevant KV vectors during decoding~\cite{zhang2023h2o, xiao2023sink, hu2025raas}, while others retain the full KV cache~\cite{tang2024quest, chen2024arkvale}, and therefore cannot reduce the memory footprint.

Although sparse attention algorithms appear beneficial, we find that they often increase \textbf{end-to-end} time and memory complexity due to frequent loss and recomputation of information, a problem that we term Lil (Less-Is-Less)\footnote{Lil abbreviates ``little'' (dropping ``tt'' and ``e''). It also suggests that sparse-attention algorithms yield ``little'' benefit. Pronunciation resembles ``Leo.''}. First, sparse attention increases \textbf{end-to-end} time complexity. Although each decode step becomes faster, the information lost during sparse attention forces the model to generate longer sequences to compensate. Second, sparse attention increases \textbf{end-to-end} memory complexity. Although each step may store fewer KV vectors, the extended generation process increases memory residency time, negating potential savings.

This paper identifies and mitigates the Lil problem, the ``elephant in the room'' of sparse attention research, in three steps. First, through a systematic empirical study, we demonstrate that widely used sparse-attention algorithms consistently increase output length by up to 90\% on reasoning-intensive datasets compared with full attention. The outputs exhibit a clear pattern of information loss followed by attempted reconstruction (Section~\ref{sec-study}). Second, we further analyze the outputs through the lens of information theory (Section~\ref{sec-theory}). We establish a quantitative relationship between the information of a sentence and its compression ratio. We further observe that, under sparse attention, the information of generated sequences does not necessarily increase as generation proceeds. Third, motivated by these empirical and theoretical findings, we propose \algo, an early-stopping algorithm that halts decoding when the information of the generated sequence ceases to increase (Section~\ref{sec-algo}). 


We implement a unified framework that supports the integration and comparative evaluation of diverse sparse-attention algorithms. We also incorporate \algo\ into this framework. Our evaluation yields two key findings (Section~\ref{sec-eval}). First, on reasoning-intensive benchmarks~\cite{math500, aime, gsm8k}, \algo\ reduces total token wastage by up to 90\% compared to decoding without early stopping, with less than 2\% accuracy drop. Second, experiments show that \algo\ can also be applied to general cases of prolonged Chain-of-Thought (CoT) generation\footnote{Prolonged CoT arises from ill reasoning patterns induced by data quality issues, human preference biases for long sentences, or reward hacking, rather than from the information-loss-and-reconstruction characteristic of the Lil problem.}.

In summary, this paper makes the following three main contributions: 
\begin{itemize}
    \item We identify and characterize the Lil problem in existing sparse-attention algorithms through a systematic empirical study.
    \item We establish a connection between the information of a sentence and its compression ratio using entropy-based compression algorithms.
    \item We propose \algo, an early-stopping algorithm (for the decode stage) that reduces token usage by up to 90\% with less than 2\% accuracy drop on reasoning-intensive benchmarks.
\end{itemize}

\section{Background and Motivation}
\label{sec-background}
\begin{figure*}[t]

\begin{center}
\centerline{\includegraphics[width=\linewidth]{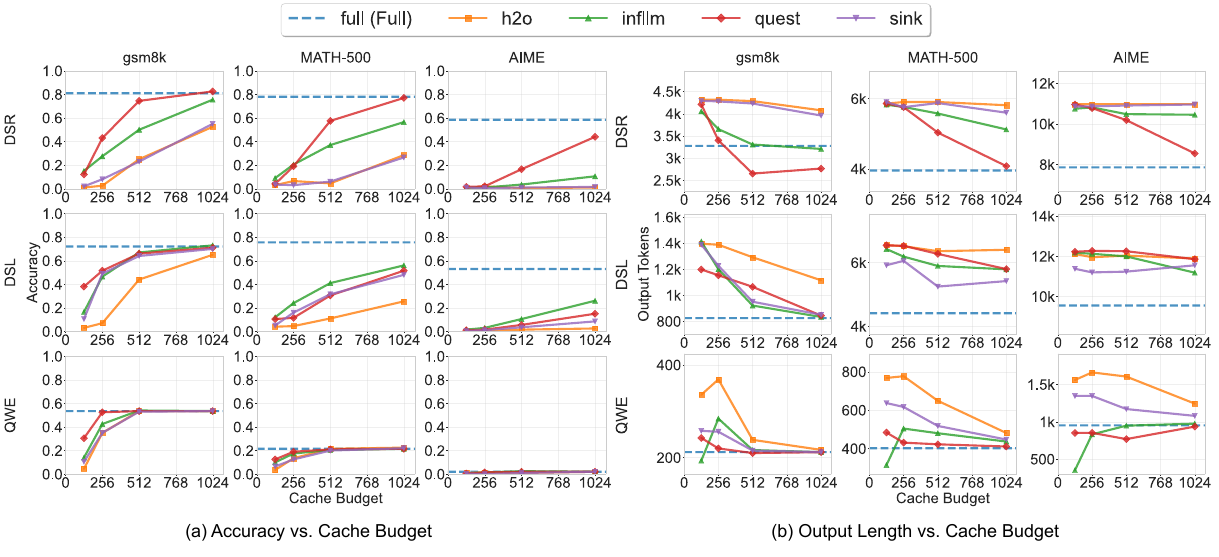}}
\caption{Accuracy/output length vs. cache budget for five algorithms (legends) across three datasets (columns) and three models (rows). DSR, DSL, and Qwe denote DeepScaleR-1.5B-Preview, DeepSeek-R1-Distill-Llama-8B, and Qwen1.5-MoE-A2.7B-Chat, respectively. The x-axis shows varying cache budgets. In (a), the y-axis shows the proportion of correctly solved problems over 200 test cases. In (b), the y-axis shows the average output length over the same 200 test cases. \textbf{For sparse-attention algorithms, the maximum generation length is capped at twice that of the full-attention baseline to prevent non-terminating generation.}}
\label{fig-study-accuracy-length}
\end{center}
\vskip -0.3in
\end{figure*}
This section reviews the LLM inference pipeline and existing sparse-attention algorithms, and highlights the key challenges that motivate our study.

\subsection{Autoregressive Generation and KV Cache}

LLMs generate tokens autoregressively, involving two stages: the prefill stage and the decode stage. In the \textbf{prefill} stage, LLMs process the entire user-provided input (prompt or a sequence of tokens) $(x_1, x_2, \dots, x_n)$. LLMs compute the Key (K) and Value (V) vectors for all tokens, store these vectors in the KV cache~\cite{hu2025epic, liu2026cacheslide}, and generate the first output token to initiate the decode stage. The prefill stage can be slow for long inputs, and the time to generate the first token is measured by the Time-to-First-Token (TTFT) metric. In the \textbf{decode} stage, LLMs generate one token at a time. The model computes the probability of the next token $x_{n+1}$, selects the most likely token, and appends its key and value vectors to the KV cache. This process repeats until a specified stopping criterion is met. The latency between consecutive tokens is measured by the Time-Between-Tokens (TBT) metric.

\subsection{Cost of Long-Prefill and Long-Decode Inference}

Both long-prefill and long-decode inference incur substantial memory and time overheads~\cite{hu2024memserve, hu2025deepserve}. In terms of \textbf{memory}, processing 128k tokens with the LLaMA~3.1~8B model in FP16 requires up to 16~GB, in addition to the 16~GB of model parameters\footnote{\url{https://huggingface.co/blog/llama31}}. In terms of \textbf{time}, inference on 32k tokens can take from tens to thousands of seconds on vLLM~0.6.1 with the same model~\cite{hu2025raas}, with large variance driven primarily by the decode stage: longer generations incur proportionally higher latency.

The decode stage dominates end-to-end inference time, especially for reasoning-intensive tasks~\cite{openai2024o1, wang2024openr, zhao2024marco, wei2022chain}. For instance, the OpenAI o1~\cite{openai2024o1} may spend tens to hundreds of seconds in internal ``thinking'' before producing a final answer\footnote{\url{https://www.reddit.com/r/OpenAI/comments/1frdwqk/your_longest_thinking_time_gpt4_o1_o1mini/}}. Accordingly, this paper focuses on optimizing long-decode inference.

\subsection{Post-Training Sparse-Attention Algorithms for Long-Decode Optimization}

To reduce memory and time complexity in LLM inference, a substantial body of work proposes sparse-attention algorithms~\cite{xiao2023sink, zhang2023h2o, tang2024quest, chen2024arkvale, nsa2025yuan}, which restrict attention computation at each decode step to a small subset of critical tokens, often comprising fewer than 10\% of the full context~\cite{tang2024quest}. First, sparse attention reduces time complexity. Full attention requires each decode token to attend to all previous tokens. In contrast, sparse attention requires each decode token to attend to only the top-k most relevant tokens. Second, sparse attention may reduce memory complexity. Some algorithms reduce memory by discarding irrelevant KV vectors during decoding, such as H$_2$O and Sink~\cite{zhang2023h2o, xiao2023sink, hu2025raas}, while others retain the full KV cache, such as infLLM and Quest~\cite{tang2024quest, chen2024arkvale, infllm2024xiao}, and therefore cannot reduce the memory footprint.

Sparse-attention algorithms can be broadly categorized into two types. \textbf{Training-aware} algorithms incorporate sparsity directly into the model architecture and training procedure. Despite being effective, they require architectural modifications and incur substantial training costs, and their sparsity is difficult to disable once deployed. Representative examples include DeepSeek Native Sparse Attention (NSA) and DeepSeek Sparse Attention (DSA)~\cite{damai2024deepseek, liu2024deepseek-v3}. In contrast, \textbf{post-training} algorithms apply sparsity to fully trained dense models at inference time. These algorithms are plug-and-play and training-free, and reportedly have demonstrated strong effectiveness in preserving accuracy while reducing latency and memory consumption. This paper focuses on Post-Training Sparse-attention algorithms in the Decode stage (PTSD).

\subsection{Limitations of PTSD}

Although sparse attention algorithms appear beneficial, we find that they often increase end-to-end time and memory complexity due to frequent loss and recomputation of information. First, sparse attention increases end-to-end time complexity. Although sparsity reduces the per-step TBT, the loss of contextual information frequently forces the model to generate longer outputs to compensate. The resulting Job Completion Time (JCT),
\begin{equation}\text{JCT}\uparrow = \text{TTFT} + \text{decode\_length}\boldsymbol{\scalebox{1.5}{$\uparrow$}} \times \text{TBT}\downarrow,
\end{equation}
increases (Section~\ref{sec-study}). Second, sparse attention increases end-to-end memory complexity. Although each decode step may store fewer KV vectors, the extended generation process increases memory residency time, negating potential savings.

We refer to this length-increasing problem as Lil (Less-Is-Less), which is the ``elephant in the room'' for the PTSD community. If left unaddressed, it undermines the fundamental motivation for adopting sparse-attention algorithms. In the subsequent sections, we first analyze the causes of this problem and then propose algorithms to mitigate it.

\section{Empirical Study of Lil}
\label{sec-study}

We evaluate five sparse-attention algorithms across three datasets and three models (see Section~\ref{sec-eval} for the detailed experimental setup), and obtain two key findings (Figure~\ref{fig-study-accuracy-length}). First, accuracy increases with cache budget. H$_2$O and Sink are less accurate under fixed budgets because they discard KV vectors. Once important information is removed, the model cannot recover it. Quest and infLLM keep all KV vectors. They maintain higher accuracy but use more memory. Second, output length decreases as the cache budget increases, but it remains longer than Full attention (by up to 90\%). With small cache budgets, information loss outpaces information gain, and the model repeats content when trying to rebuild context (Figure~\ref{fig-study-examples} (a)). The model may fail to solve the task and
generate indefinitely due to lost context. With larger cache budgets, the model makes partial progress and solves more cases; however, the outputs may still be excessively long, because the correct answer is produced early but the model continues verification and subsequently forgets that it has already generated the answer (Figure~\ref{fig-study-examples} (b)).

\section{Compression Theory}
\label{sec-theory}

\begin{figure}[t]

\begin{center}
\centerline{\includegraphics[width=\linewidth]{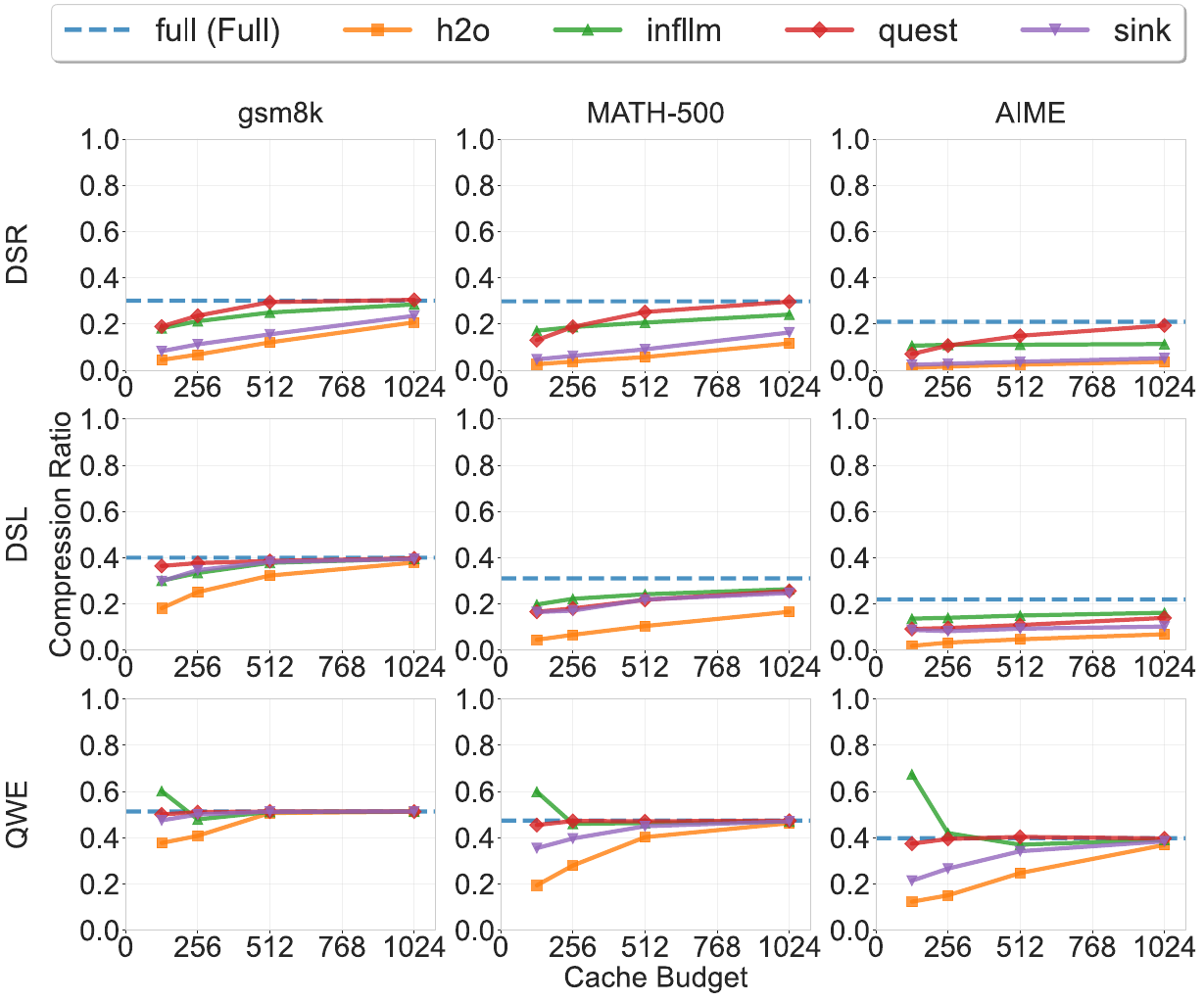}}
\caption{Compression ratio vs. cache budget. Notations follow Figure~\ref{fig-study-accuracy-length}. The y-axis shows the average compression ratio (compressed-sequence length / original-sequence length) over 200 test cases.}
\label{fig-theory-compression-eval}
\end{center}
\vskip -0.3in
\end{figure}

Section~\ref{sec-study} shows the Lil problem: information is lost under sparse attention, and models attempt to regain it by generating more tokens. We next use information theory to analyze this loss and gain.

To this end, we adopt LZ77~\cite{ziv1977lz77}, a compression algorithm that is simple, efficient, and grounded in theory. First, the key insight of LZ77 is simple: replacing repeated substrings with concise references (e.g., offset-length pairs) to their earlier occurrences. When later segments of a sequence frequently recur in earlier segments, the resulting compression ratio (compressed length divided by original length) is small. Second, LZ77 is computationally efficient: compressing a sequence of 128k tokens takes approximately 34 ms (Section~\ref{sec-eval}), which is comparable to the time to decode a single token~\cite{zheng2024sglang, xds2025xds}. Third, LZ77 admits strong theoretical guarantees. The achieved compression ratio $\rho$ satisfies
\begin{equation}
\rho - \epsilon(L_s) \le h(L_s - 1) \le \rho.
\end{equation}
where $h(k)$ denotes per‑symbol entropy, and $\epsilon(L_s) = \mathcal{O}(\log L_s / L_s)$.
Consequently, $\rho$ estimates information entropy up to a small term. A lower value of $\rho$ indicates less new information and more redundancy. Please refer to the appendix for a comprehensive illustration of the LZ77 algorithm and related proof. 

We compress all sequences in Figure~\ref{fig-study-accuracy-length} and report their corresponding compression ratios in Figure~\ref{fig-theory-compression-eval}, from which we draw two key insights. First, despite producing longer outputs, sparse-attention algorithms generate sequences with substantially less information than full attention. This result indicates that the model largely repeats earlier content to reconstruct lost information, leading LZ77 to encode much of the later sequence as references to earlier segments. Second, as the cache budget increases, information gain increasingly outpaces information loss. As a result, the model attains higher accuracy with fewer tokens, diminishing the need for information reconstruction. Correspondingly, the compression ratio approaches that of full attention, reflecting more informative and less redundant generation.

\section{Early-Stopping Algorithm}
\label{sec-algo}

\begin{algorithm}[t]
\caption{\algo\ Algorithm}\label{algo}
\begin{algorithmic}[1]
\STATE \textbf{Input:} A sequence X of prefill tokens, a model M, a frequency $f$, and a threshold $t$
\STATE \textbf{Output:} A sequence Y of prefill tokens plus decode tokens
\STATE
\STATE cnt = 1
\STATE lastCompress = LZ77(X)
\STATE curCompress = lastCompress
\STATE
\STATE Y = X
\STATE y = M.forward(Y, ``prefill'')
\STATE \textbf{while} y $\neq$ eos and len(Y) < M.context\_len() and not is\_early\_stop(Y)
    \STATE\hspace{0.5cm}Y.append(y)
    \STATE\hspace{0.5cm}y = M.forward(Y, ``decode'')
\STATE \textbf{Return} Y

\STATE 

\STATE \textbf{Function} is\_early\_stop(Y)
\STATE\textbf{if} cnt \% f == 0
    \STATE\hspace{0.5cm}curCompress = LZ77(Y)
    \STATE\hspace{0.5cm}\textbf{if} curCompress - lastCompress < t
        \STATE\hspace{1cm}\textbf{Return} True
    \STATE\hspace{0.5cm} lastCompress = curCompress
\STATE cnt = cnt + 1
\STATE \textbf{Return} False
\STATE \textbf{End Function}
\end{algorithmic}
\end{algorithm}
\begin{figure*}[t]

\begin{center}
\centerline{\includegraphics[width=\linewidth]{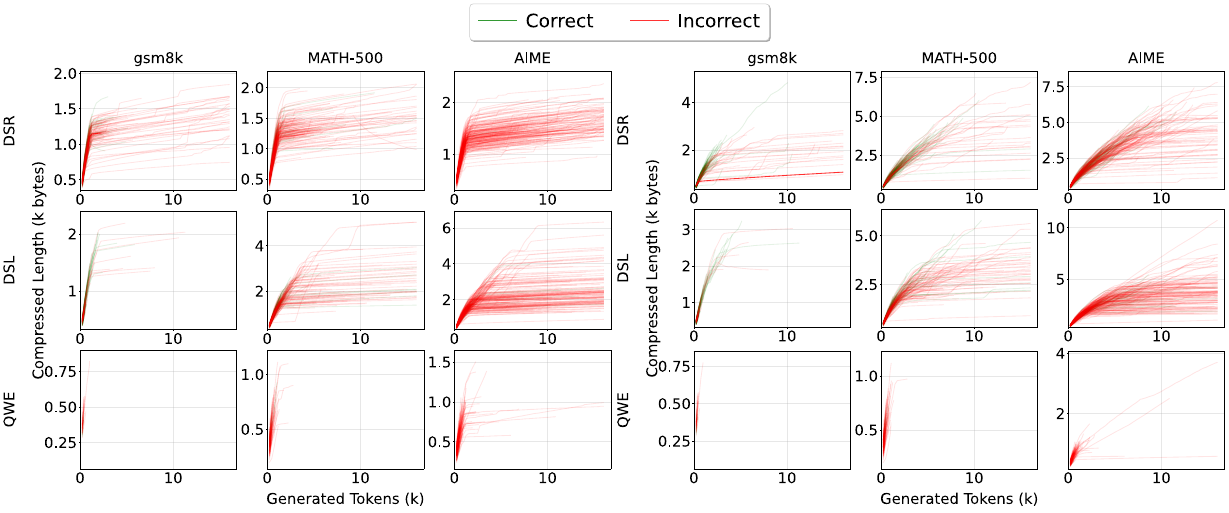}}
\caption{Compressed length (y-axis) vs. original length (x-axis) for Sink with 1024 cache budget (left) and Quest with 1024 cache budget (right) across three models and three datasets. Each line indicates an individual test case. Other notations follow Figure~\ref{fig-study-accuracy-length}. As the model generates more tokens (i.e., as the original length increases), the compressed length initially grows rapidly and then plateaus. Green curves denote correct test cases, whereas red curves denote incorrect test cases.}
\label{fig-algo-delta}
\end{center}
\vskip -0.3in
\end{figure*}

Based on the preceding empirical and theoretical analyses and Figure~\ref{fig-algo-delta}, we observe that under sparse-attention algorithms, the model may cease to gain new information while continuing to generate tokens. Motivated by this insight, we propose \algo, an early-stopping algorithm that detects sustained stagnation in information gain during generation and terminates decoding early to reduce token consumption.

The algorithm is shown in Algorithm~\ref{algo}. First, in the prefill stage, the user prompt is processed and compressed to obtain \textit{lastCompress}, the number of bytes left after compression. Second, the model enters the decode stage and stops upon generating an \textit{eos} token or reaching the maximum context length. \algo\ introduces an additional stopping condition: every $f$ decode steps, the current sequence (prefill plus generated tokens) is compressed to obtain \textit{curCompress} and compared with \textit{lastCompress}. If the increase is below a threshold $t$ bytes, the sequence has gained negligible new information; that is, the newly generated tokens are largely redundant under LZ77, and the generation is therefore terminated.

We next discuss the choice of $f$ and $t$. First, the interval $f$ must balance computational overhead and responsiveness: a small $f$ incurs frequent compression and excessive overhead, whereas a large $f$ reduces the sensitivity of \algo\ and delays termination, thereby diminishing token savings. We set $f = 250$ in our experiments. In practice, $f$ primarily affects token savings and end-to-end latency, but not accuracy, and can be adjusted based on model architectures, sequence lengths, and GPUs, which determine the per-step decoding cost. Second, given a fixed $f$, the threshold $t$ is chosen such that $t / f$ lies between the slopes of the initial growth stage and the subsequent plateau stage in Figure~\ref{fig-algo-delta}. Empirically, the slope in the plateau stage is below $0.02$ across datasets and models, whereas the initial slope exceeds $1$, and we set $t = 20$ such that $t / f \approx 0.08$. Because the transition between the two stages is sharp, \algo\ is robust to variations in $t$ as long as $t / f$ falls within this range.

\section{Evaluation}
\label{sec-eval}

\begin{table*}[t]
\centering
\caption{Average token savings and accuracy degradation after applying \algo. Notations follow Figure~\ref{fig-study-accuracy-length}. Cell entries follow the format $Savings / \Delta Accuracy$, representing the percentage of relative token savings and the corresponding relative accuracy shift. All values are expressed in \%. Please refer to Figure~\ref{fig-study-accuracy-length} for absolute token and accuracy numbers.}
\small
\setlength{\tabcolsep}{1.5pt}
\resizebox{\linewidth}{!}{%
\begin{tabular}{cc r@{/}l r@{/}l r@{/}l r@{/}l r@{/}l r@{/}l r@{/}l r@{/}l r@{/}l r@{/}l r@{/}l r@{/}l}
\toprule
 &  & \multicolumn{8}{c}{\textbf{GSM8K}} 
    & \multicolumn{8}{c}{\textbf{MATH-500}}
    & \multicolumn{8}{c}{\textbf{AIME}} \\
\cmidrule(lr){3-10} \cmidrule(lr){11-18} \cmidrule(lr){19-26}
 & \% & \multicolumn{2}{c}{H2O} & \multicolumn{2}{c}{Sink} & \multicolumn{2}{c}{Quest} & \multicolumn{2}{c}{infLLM} & \multicolumn{2}{c}{H2O} & \multicolumn{2}{c}{Sink} & \multicolumn{2}{c}{Quest} & \multicolumn{2}{c}{infLLM} & \multicolumn{2}{c}{H2O} & \multicolumn{2}{c}{Sink} & \multicolumn{2}{c}{Quest} & \multicolumn{2}{c}{infLLM} \\
\midrule
\multirow{4}{*}{\rotatebox{90}{DSR}} & 128 & 67.8 & 0.0 & 68.1 & 0.0 & 6.7 & 0.0 & 6.4 & 0.0 & 81.8 & 0.0 & 84.1 & \textcolor{green}{+1.5} & 11.5 & 0.0 & 8.3 & 0.0 & 93.7 & 0.0 & 92.5 & 0.0 & 18.4 & \textcolor{red}{-0.5} & 14.1 & 0.0 \\
 & 256 & 62.1 & 0.0 & 64.9 & \textcolor{green}{+0.5} & 10.2 & 0.0 & 5.4 & \textcolor{red}{-0.5} & 78.0 & \textcolor{red}{-0.5} & 79.9 & \textcolor{green}{+1.0} & 11.2 & 0.0 & 9.0 & \textcolor{red}{-0.5} & 91.2 & 0.0 & 90.9 & \textcolor{red}{-0.5} & 19.7 & \textcolor{red}{-0.5} & 16.8 & \textcolor{green}{+0.5} \\
 & 512 & 49.3 & 0.0 & 53.9 & \textcolor{green}{+1.0} & 6.1 & \textcolor{red}{-1.0} & 6.4 & \textcolor{red}{-0.5} & 66.8 & 0.0 & 70.9 & \textcolor{red}{-0.5} & 8.6 & 0.0 & 10.1 & \textcolor{green}{+0.5} & 86.4 & 0.0 & 87.8 & \textcolor{red}{-0.5} & 20.3 & 0.0 & 24.1 & 0.0 \\
 & 1024 & 36.9 & 0.0 & 32.8 & \textcolor{red}{-1.5} & 9.8 & \textcolor{red}{-2.5} & 10.9 & \textcolor{red}{-1.5} & 53.6 & 0.0 & 51.8 & \textcolor{red}{-0.5} & 6.0 & \textcolor{red}{-0.5} & 11.9 & 0.0 & 79.6 & 0.0 & 79.9 & 0.0 & 14.5 & \textcolor{red}{-1.0} & 33.2 & \textcolor{green}{+0.5} \\
\midrule
\multirow{4}{*}{\rotatebox{90}{DSL}} & 128 & 22.7 & 0.0 & 8.1 & 0.0 & 1.6 & 0.0 & 0.4 & 0.0 & 75.5 & 0.0 & 47.9 & \textcolor{green}{+1.0} & 15.1 & \textcolor{green}{+1.0} & 3.8 & \textcolor{green}{+0.5} & 91.0 & 0.0 & 71.0 & 0.0 & 23.2 & \textcolor{green}{+0.5} & 4.8 & 0.0 \\
 & 256 & 23.4 & 0.0 & 10.2 & 0.0 & 3.6 & \textcolor{red}{-0.5} & 0.7 & 0.0 & 75.8 & 0.0 & 45.8 & \textcolor{red}{-0.5} & 23.9 & \textcolor{green}{+2.5} & 1.7 & 0.0 & 87.9 & 0.0 & 72.6 & 0.0 & 44.9 & 0.0 & 3.9 & 0.0 \\
 & 512 & 18.5 & 0.0 & 6.4 & \textcolor{green}{+1.0} & 2.4 & \textcolor{green}{+0.5} & 0.3 & \textcolor{red}{-0.5} & 71.0 & \textcolor{green}{+0.5} & 35.7 & \textcolor{red}{-1.0} & 26.3 & \textcolor{red}{-1.0} & 3.9 & \textcolor{red}{-0.5} & 87.4 & 0.0 & 67.6 & 0.0 & 53.9 & 0.0 & 10.0 & \textcolor{red}{-0.5} \\
 & 1024 & 9.6 & 0.0 & 3.6 & \textcolor{red}{-0.5} & 1.5 & \textcolor{red}{-0.5} & 0.3 & 0.0 & 58.8 & 0.0 & 29.9 & 0.0 & 22.3 & \textcolor{green}{+2.0} & 5.9 & 0.0 & 83.6 & 0.0 & 65.0 & 0.0 & 48.2 & \textcolor{red}{-1.0} & 16.0 & \textcolor{red}{-2.0} \\
\midrule
\multirow{4}{*}{\rotatebox{90}{Qwe}} & 128 & 0.9 & 0.0 & 1.1 & 0.0 & 0.0 & 0.0 & 0.0 & 0.0 & 16.7 & 0.0 & 14.2 & \textcolor{red}{-0.5} & 0.2 & 0.0 & 0.0 & 0.0 & 39.5 & 0.0 & 36.8 & \textcolor{green}{+0.5} & 1.4 & 0.0 & 0.2 & 0.0 \\
 & 256 & 0.5 & 0.0 & 1.7 & 0.0 & 0.0 & 0.0 & 0.0 & 0.0 & 19.3 & 0.0 & 13.7 & 0.0 & 0.0 & 0.0 & 0.4 & 0.0 & 42.7 & 0.0 & 31.7 & 0.0 & 1.1 & 0.0 & 0.8 & 0.0 \\
 & 512 & 1.3 & 0.0 & 0.1 & 0.0 & 0.0 & 0.0 & 0.0 & 0.0 & 12.5 & \textcolor{green}{+0.5} & 6.0 & 0.0 & 0.1 & 0.0 & 0.4 & 0.0 & 31.6 & 0.0 & 16.1 & 0.0 & 0.6 & 0.0 & 1.1 & 0.0 \\
 & 1024 & 0.2 & 0.0 & 0.0 & 0.0 & 0.0 & 0.0 & 0.0 & 0.0 & 3.2 & 0.0 & 0.8 & 0.0 & 0.2 & 0.0 & 0.6 & 0.0 & 11.0 & 0.0 & 6.0 & 0.0 & 1.6 & \textcolor{green}{+0.5} & 2.7 & 0.0 \\
\bottomrule
\end{tabular}%
}
\label{tab-eval-e2e}
\vskip -0.2in
\end{table*}

In this section, we begin by describing the experimental setup, including implementation details, datasets, models, evaluation metrics,
and the software/hardware environment. We then present key evaluation results.

\subsection{Experimental Setup}

\textbf{Implementation.} We implement \algo\ based on Hugging Face~\cite{hf} with 2k lines of Python code. We port the LZ77 algorithm from the widely used compression tool Gzip\footnote{https://www.gzip.org/}.

\textbf{Datasets.} We take the first 200 test cases from each of the following three open-source datasets as our benchmarks: GSM8K~\cite{gsm8k}, MATH500~\cite{math500}, and AIME~\cite{aime}, to test the reasoning ability of language models. First, GSM8K~\cite{gsm8k} contains 8.5k high-quality, linguistically diverse grade-school math problems. These human-written problems require solutions that involve multi-step reasoning and a series of basic arithmetic operations. Second, MATH500~\cite{math500} contains 500 challenging problems sourced from high-school math competitions with five distinct levels based on the Art of Problem Solving (AoPS) framework, ranging from level 1 to level 5. Third, AIME~\cite{aime} is a math-problem dataset collected from the American Invitational Mathematics Examination (AIME) competition from 1983 to 2024, designed to challenge the most exceptional high-school math students in the United States. These problems cover various fields, such as algebra, geometry, and number theory.

\textbf{Models.} We evaluate our algorithm using three popular models: DeepScaleR-1.5B-Preview\footnote{https://pretty-radio-b75.notion.site/DeepScaleR-Surpassing-O1-Preview-with-a-1-5B-Model-by-Scaling-RL-19681902c1468005bed8ca303013a4e2}, DeepSeek-R1-Distill-Llama-8B~\cite{deepseekai2025deepseekr1}, and Qwen1.5-MoE-A2.7B-Chat\footnote{https://qwenlm.github.io/blog/qwen-moe/}. These models span diverse architectures (dense vs. MoE), training recipes, and parameter scales. In this paper, we use DSR to denote DeepScaleR-1.5B-Preview, DSL to denote DeepSeek-R1-Distill-Llama-8B, and Qwe to denote Qwen1.5-MoE-A2.7B-Chat.

\textbf{Metrics.} We evaluate efficiency and accuracy using two metrics. \emph{Token savings} is defined as the ratio of the number of tokens generated without \algo\ to that generated after applying \algo. \emph{Accuracy}~\cite{wang2024openr} measures the mathematical equivalence between an LLM's output and the ground-truth answer. For each test case, it is either correct or incorrect, and the overall accuracy is reported as the percentage of correctly solved problems across the entire dataset. When reporting accuracy changes, we use the absolute difference in accuracy, expressed in percentage points.

\textbf{Baselines.} We compare Full, H$_2$O, StreamingLLM, InfLLM, and Quest (spanning sparse-attention algorithms from the earliest proposals to the state of the art as of October 2025) with and without \algo. For each algorithm, the cache budget denotes the maximum number of tokens to which a decoding token can attend. For H$_2$O and Sink, the cache budget additionally specifies the number of tokens whose KVs are retained, as these algorithms attend to only preserved tokens. In contrast, infLLM and Quest retain the KVs of all tokens.

\textbf{Environment.} We run experiments on a server with a single NVIDIA A100-80GB GPU. The server has a 128-core Intel(R) Xeon(R) Platinum 8358P CPU@2.60GHz with two hyperthreads and 1~TB DRAM. We use Ubuntu 20.04 with Linux kernel 5.16.7 and CUDA 12.6.

\subsection{Evaluation Results}

\textbf{\algo\ saves up to 90\% tokens with less than 2\% accuracy drop.} Table~\ref{tab-eval-e2e} reports the end-to-end results after applying \algo, from which we draw four key findings. First, \algo\ reduces token usage by up to 90\% while incurring less than a 2\% accuracy drop, demonstrating its effectiveness. Second, in some cases, \algo\ even improves accuracy. This improvement occurs when the model generates the correct answer early but continues decoding, redundantly re-evaluating the solution, and ultimately loses the correct answer. Early termination prevents such degradation. Third, as the sparsity algorithm becomes more conservative (e.g., from Sink to Quest, or from a cache budget of 128 to 1024), token redundancy decreases, resulting in fewer tokens saved. Nevertheless, even the most conservative sparse configurations that achieve accuracy comparable to full attention still exhibit measurable redundancy and benefit from early stopping (e.g., the DSR model on GSM8K with Quest and a 1024 cache budget). Fourth, \algo\ yields negligible token savings for the Qwen MoE model that is not specialized for mathematical reasoning and tends to generate short, incorrect responses (Figure~\ref{fig-study-accuracy-length}). For models with limited capability and short chains of thought, such as Qwen MoE, early stopping provides limited benefit.

\textbf{Cost of LZ77.} We generate random strings of up to 128k tokens and measure the cost of LZ77 compression, which is approximately 34 ms. Because random strings exhibit minimal repetition and are more expensive to compress than natural language text, and because 128k tokens far exceed the sequence lengths used in our experiments, this measurement represents an upper bound. The resulting cost is comparable to the latency of decoding a single token~\cite{zheng2024sglang}. With $f = 250$, compression is invoked once every 250 decoding steps, making the overall overhead negligible.

\textbf{Token savings on correct and incorrect cases.} Table~\ref{tab-eval-correct-incorrect} reports token savings separately for correct and incorrect test cases, revealing two key observations. First, \algo\ derives most of its effectiveness from terminating incorrect generations that would otherwise continue indefinitely; to avoid overstating this effect, we truncate such generations at twice the length of sequences produced under full attention (see the caption of Figure~\ref{fig-study-accuracy-length}). Second, \algo\ also reduces tokens in correct cases where the model has already produced the correct answer but continues to generate redundant reasoning, repeatedly rechecking the solution due to loss of earlier context.

\begin{table*}[t]
\centering
\caption{Average token savings for correct and incorrect cases. Notation follows Table~\ref{tab-eval-e2e}. Each cell reports (token savings on correct cases) / (token savings on incorrect cases).}
\small
\setlength{\tabcolsep}{1.5pt}
\resizebox{\linewidth}{!}{%
\begin{tabular}{cc r@{/}l r@{/}l r@{/}l r@{/}l r@{/}l r@{/}l r@{/}l r@{/}l r@{/}l r@{/}l r@{/}l r@{/}l}
\toprule
& & \multicolumn{8}{c}{\textbf{GSM8K}}
  & \multicolumn{8}{c}{\textbf{MATH-500}}
  & \multicolumn{8}{c}{\textbf{AIME}} \\
\cmidrule(lr){3-10} \cmidrule(lr){11-18} \cmidrule(lr){19-26}
 & \% & \multicolumn{2}{c}{H2O} & \multicolumn{2}{c}{Sink} & \multicolumn{2}{c}{Quest} & \multicolumn{2}{c}{infLLM} & \multicolumn{2}{c}{H2O} & \multicolumn{2}{c}{Sink} & \multicolumn{2}{c}{Quest} & \multicolumn{2}{c}{infLLM} & \multicolumn{2}{c}{H2O} & \multicolumn{2}{c}{Sink} & \multicolumn{2}{c}{Quest} & \multicolumn{2}{c}{infLLM} \\
\midrule
\multirow{4}{*}{\rotatebox{90}{DSR}} & 128 & 83.0 & 67.7 & 59.3 & 68.3 & 0.9 & 7.7 & 3.8 & 6.8 & 77.4 & 82.0 & 79.7 & 84.3 & 7.3 & 11.7 & 3.0 & 8.8 & 89.1 & 93.7 & 84.8 & 92.6 & 37.2 & 18.2 & 17.6 & 14.1 \\
 & 256 & 55.9 & 62.3 & 27.0 & 68.4 & 1.5 & 18.2 & 1.1 & 7.1 & 68.9 & 78.7 & 72.5 & 80.2 & 3.4 & 13.7 & 1.2 & 11.1 & 92.3 & 91.2 & 0.0 & 90.9 & 0.0 & 20.0 & 28.0 & 16.6 \\
 & 512 & 23.7 & 57.9 & 20.1 & 64.9 & 0.3 & 25.7 & 0.0 & 13.2 & 52.0 & 67.5 & 32.0 & 73.6 & 2.1 & 18.4 & 3.1 & 14.3 & 78.1 & 86.5 & 90.6 & 87.8 & 1.2 & 24.5 & 5.2 & 25.0 \\
 & 1024 & 15.1 & 61.9 & 6.8 & 63.9 & 0.0 & 50.4 & 0.5 & 41.4 & 26.6 & 64.8 & 16.7 & 65.7 & 0.9 & 23.6 & 1.0 & 26.2 & 49.8 & 79.9 & 30.7 & 80.6 & 0.6 & 25.0 & 5.0 & 36.9 \\
\midrule
\multirow{4}{*}{\rotatebox{90}{DSL}} & 128 & 21.6 & 22.8 & 3.8 & 8.6 & 0.0 & 2.6 & 0.0 & 0.5 & 83.9 & 75.2 & 41.5 & 48.3 & 19.8 & 14.3 & 1.9 & 4.0 & 91.9 & 91.0 & 59.2 & 71.0 & 30.7 & 23.1 & 0.0 & 4.9 \\
 & 256 & 17.8 & 23.8 & 0.6 & 19.5 & 0.3 & 7.8 & 0.0 & 1.4 & 68.8 & 76.1 & 16.6 & 51.5 & 12.4 & 26.3 & 0.0 & 2.4 & 96.9 & 87.8 & 88.3 & 72.4 & 21.2 & 45.7 & 0.0 & 4.1 \\
 & 512 & 4.9 & 29.5 & 1.1 & 16.4 & 0.7 & 6.0 & 0.0 & 0.9 & 46.5 & 74.2 & 7.7 & 49.7 & 10.4 & 35.0 & 0.1 & 6.8 & 73.5 & 87.6 & 22.0 & 69.3 & 33.8 & 55.8 & 6.7 & 10.4 \\
 & 1024 & 3.7 & 20.5 & 0.8 & 10.0 & 0.4 & 4.0 & 0.0 & 1.0 & 21.0 & 71.7 & 8.2 & 52.0 & 7.4 & 42.5 & 1.1 & 12.2 & 60.3 & 84.2 & 23.4 & 69.3 & 21.6 & 54.2 & 2.3 & 20.7 \\
\midrule
\multirow{4}{*}{\rotatebox{90}{Qwe}} & 128 & 0.0 & 0.9 & 0.0 & 1.2 & 0.0 & 0.0 & 0.0 & 0.0 & 9.6 & 17.0 & 7.7 & 14.6 & 0.0 & 0.2 & 0.0 & 0.0 & 66.3 & 39.3 & 66.4 & 36.5 & 0.0 & 1.4 & 0.0 & 0.2 \\
 & 256 & 0.0 & 0.8 & 0.0 & 2.5 & 0.0 & 0.0 & 0.0 & 0.0 & 7.0 & 21.2 & 2.0 & 15.4 & 0.0 & 0.0 & 0.0 & 0.5 & 48.6 & 42.6 & 0.0 & 31.7 & 0.0 & 1.2 & 0.0 & 0.8 \\
 & 512 & 0.0 & 2.8 & 0.0 & 0.3 & 0.0 & 0.0 & 0.0 & 0.0 & 6.5 & 14.2 & 0.0 & 7.5 & 0.0 & 0.2 & 0.0 & 0.5 & 0.0 & 32.0 & 0.0 & 16.2 & 0.0 & 0.6 & 0.0 & 1.1 \\
 & 1024 & 0.0 & 0.5 & 0.0 & 0.0 & 0.0 & 0.0 & 0.0 & 0.0 & 2.0 & 3.6 & 0.0 & 1.0 & 0.0 & 0.2 & 0.9 & 0.5 & 0.0 & 11.2 & 0.0 & 6.2 & 9.0 & 1.4 & 0.0 & 2.7 \\
\bottomrule
\end{tabular}%
}
\label{tab-eval-correct-incorrect}
\vskip -0.2in
\end{table*}

\textbf{Token savings on full attention.} 
\begin{table}[t]
\centering
\caption{Average token savings and accuracy degradation after applying \algo\ on the full attention algorithm. Notation follows Table~\ref{tab-eval-e2e}.}
\small
\setlength{\tabcolsep}{8pt}
\begin{tabular}{l r@{/}l r@{/}l r@{/}l}
\toprule
 & \multicolumn{2}{c}{\textbf{GSM8K}} & \multicolumn{2}{c}{\textbf{MATH-500}} & \multicolumn{2}{c}{\textbf{AIME}} \\
\midrule
DSR & 12.5 & \textcolor{red}{-0.5} & 9.4 & \textcolor{red}{-0.5} & 18.3 & \textcolor{red}{-1.5} \\
DSL & 0.9 & \textcolor{green}{+0.5} & 5.9 & 0 & 15.4 & \textcolor{red}{-0.5} \\
Qwe & 0.0 & 0 & 0.0 & 0 & 1.3 & 0 \\
\bottomrule
\end{tabular}
\label{tab-eval-full}
\vskip -0.2in
\end{table}
Figure~\ref{fig-theory-compression-eval} shows that even sequences generated with full attention achieve low compression ratios, indicating substantial redundancy. Table~\ref{tab-eval-full} further demonstrates that the early-stopping algorithm \algo\ reduces token usage under full attention as well. This behavior is analogous to Chain-of-Thought (CoT) compression approaches that aim to mitigate ineffective reasoning patterns arising from training data artifacts, human preferences for verbose explanations, or reward hacking (see Section~\ref{sec-related}). We therefore conclude that \algo\ is applicable beyond sparse-attention settings and can be used more generally to address prolonged CoT generation. A direct comparison between \algo\ and existing CoT compression approaches is left for future work.

\section{Related Work}
\label{sec-related}

\subsection{Sparse-Attention Algorithms} 

Sparse-attention algorithms exploit the fact that only a small subset of tokens receives high attention scores, and we categorize these algorithms along two dimensions. (1) Inference stage. Some algorithms primarily accelerate the prefill stage, such as MInference~\cite{jiang2024minference}, FlexPrefill~\cite{lai2025flexprefill}, and SpargeAttention~\cite{zhang2025spargeattn}, while others focus on the decode stage, including H$_2$O~\cite{zhang2023h2o}, StreamingLLM~\cite{xiao2023sink}, Quest~\cite{tang2024quest}, InfLLM~\cite{infllm2024xiao}, DuoAttention~\cite{xiao2024duoattention}, and RaaS~\cite{hu2025raas}. (2) Training dependency. Training-aware algorithms incorporate sparsity into the model architecture and are trained jointly with the model, such as DeepSeek NSA~\cite{nsa2025yuan} and DSA~\cite{liu2024deepseek-v3}. In contrast, training-free algorithms operate as plug-ins and can be readily applied to pretrained models; examples include H$_2$O, StreamingLLM, Quest, InfLLM, DuoAttention, and RaaS.

In this paper, we focus on Post-Training Sparse attention in the Decode stage (PTSD) for three reasons. First, the decode stage largely determines model performance, particularly in long-reasoning tasks, making it a critical target for optimization. Second, post-training sparsity is plug-and-play and can be integrated into existing models without retraining, making it practical for real-world deployment. Third, abundant sparse-attention algorithms fall into the PTSD category, making it a natural and impactful focus of study.

\subsection{Chain-of-Thought Compression}

Chain-of-thought (CoT) reasoning enhances model capability by enabling step-by-step inference that incrementally connects the prompt to the final answer. However, CoT can become unnecessarily long and redundant due to suboptimal training practices, such as reinforcement learning setups that inadvertently encourage verbose reasoning through reward hacking. Consequently, a line of research has emerged to focus on compressing prolonged CoT sequences.

Prior work on CoT compression, often referred to as long-to-short reasoning, can be broadly categorized into two classes. Training-aware approaches integrate compression into the training process, including ThinkPrune~\cite{hou2025thinkprune}, DLER~\cite{liu2025dler}, and O1-Pruner~\cite{luo2025o1pruner}. In contrast, post-training approaches operate on pretrained models, such as Answer5~\cite{liu2025answer5}, HALT-CoT~\cite{laaouach2025haltcot}, and UnCerT-CoT~\cite{zhu2025uncert}.

Although the approach proposed in this paper also applies to CoT compression, the root causes of lengthy reasoning differ. First, in our setting, lengthy CoT arises from the use of sparse-attention algorithms, which induce repeated information loss and reconstruction. Second, by contrast, in long-to-short reasoning research, lengthy CoT typically stems from ill-reasoning patterns caused by data quality issues, human preference biases, or reward hacking. 

These two root causes are largely orthogonal: when CoT is already lengthy due to ill patterns, sparse attention can further exacerbate its length. Although \algo\ is designed to compress lengthy CoT induced by sparse attention, we observe that it also generalizes to ill-patterned CoT (Table~\ref{tab-eval-full}), which we leave for future exploration.
\section{Conclusion}
\label{sec-conclusion}

Sparse-attention algorithms speed up each decode step but can hurt end-to-end efficiency because information loss during sparse decoding leads to repeated generation and longer outputs. We have identified this problem as ``Lil,'' quantified redundancy with compression ratio, and proposed \algo\ to stop decoding when information loss exceeds information gain. \algo\ reduces unnecessary token generation without harming output quality. Beyond sparse decoding, \algo\ can also be applied to general cases of prolonged Chain-of-Thought (CoT) generation. In future work, we plan to study \algo\ on prolonged CoT, which often arises from flawed reasoning patterns rather than the information-loss-and-reconstruction characteristic of the Lil problem.

\section*{Acknowledgments}
This work was partially supported by the Fundamental and Interdisciplinary Disciplines Breakthrough Plan of the Ministry of Education of China (No. JYB2025XDXM118), National Natural Science Foundation of China under Grant No. U25A6023, 92464301, 625B2002, and Tencent Hunyuan Fellowship. We would also like to thank the anonymous reviewers for their insightful comments and suggestions, which help improve the quality of this paper.  

\section*{Limitations}
\label{sec-limitations}

Our work has the following major limitations. 

\textbf{Evaluation on a limited set of datasets and models.} Our evaluation covers only three models and three datasets. As such, the results may not generalize beyond these specific configurations. Although models with longer context lengths (e.g., Qwen2.5-Max, DeepSeek-r1) and datasets such as GPQA Diamond and Codeforces exist, exhaustive evaluation across all combinations is computationally prohibitive~\cite{hu2023pcrml}. As reported in prior work~\cite{zhong2024distserve}, decoding a single token can take approximately 30 ms; thus, processing 16k tokens on an A100-80GB GPU requires around 8 minutes. Running 200 test cases would take over a day on a single GPU, making large-scale evaluation infeasible with limited resources. Despite these constraints, we select datasets spanning three levels of difficulty and models covering diverse architectures (dense vs. MoE), training recipes, and parameter scales. We therefore believe that the Lil problem is not an artifact of specific configurations but is universal across datasets and models.

\textbf{Evaluation on a limited set of sparse-attention algorithms.} Our evaluation covers only four sparse-attention algorithms, and thus the results may not directly generalize beyond this set. Nevertheless, we contend that the Lil problem is inherent to sparse-attention algorithms, because all such algorithms assume sparsity patterns---i.e., that a few tokens are important---and may lose information (Section~\ref{sec-theory}). We select diverse algorithms, including the pioneering H$_2$O, the widely used StreamingLLM (recently applied in GPT-OSS~\cite{openai2025gptoss}), the state-of-the-art Quest, and its counterpart infLLM. Other algorithms, such as ClusterKV~\cite{liu2025clusterkv} and PQCache~\cite{zhang2025pqcache}, differ only in how K vectors are grouped and do not alter the use of sparsity. Hence, we believe that the Lil problem is not specific to particular configurations but is universal across sparse-attention algorithms.

\bibliography{custom}

\appendix
\begin{figure*}[t]

\begin{center}
\centerline{\includegraphics[width=\linewidth]{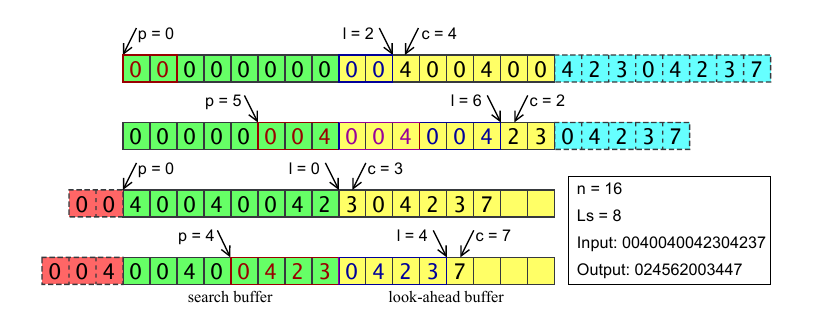}}
\caption{Illustration of the LZ77 algorithm.}
\label{fig-theory-compression-algo}
\end{center}

\end{figure*}
\begin{figure*}[t]

\begin{center}
\centerline{\includegraphics[width=\linewidth]{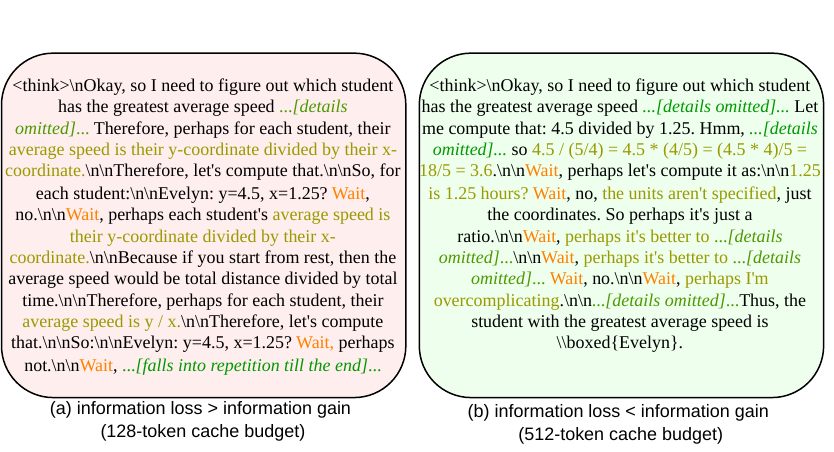}}
\caption{Examples of information loss and gain under sparse attention.}
\label{fig-study-examples}
\end{center}
\end{figure*}

\section{LZ77 Compression Algorithm}

We describe the LZ77 algorithm in detail. LZ77 operates using a sliding window buffer of fixed length \(n\), divided into two parts:
\begin{itemize}
    \item A \textit{search buffer} of length \(n - L_s\), which holds recently encoded data and acts as a dynamic dictionary.
    \item A \textit{look-ahead buffer} of length \(L_s\), which contains the subsequence awaiting encoding.
\end{itemize}

The encoding process proceeds iteratively. At each step, the algorithm finds the longest prefix of the look-ahead buffer that matches a substring within the search buffer. This match is encoded as a triple \((p, l, c)\), where
\begin{itemize}
    \item \(p\) is the offset (distance) to the start of the match in the search buffer,
    \item \(l\) is the length of the matched substring,
    \item \(c\) is the first character that does not match (the literal following the match).
\end{itemize}
After encoding, both buffers are advanced by \(l + 1\) characters, moving the processed symbols into the search buffer for potential future matches.

To concretely illustrate the encoding mechanism, Figure~\ref{fig-theory-compression-algo} provides a step-by-step visualization of LZ77 compressing the example sequence \texttt{0040040042304237}. The visualization uses distinct color-coded regions: the \textit{search buffer} (green) holds the recently processed data available for matching; the \textit{look-ahead buffer} (yellow) contains the pending sequence to be encoded; unprocessed input is shown in blue; and data that has shifted out of the search window is marked in red.

The algorithm iterates the same core loop. Below is a trace of its execution:

\begin{enumerate}
    \item The first eight characters (\texttt{00400400}) are loaded into the look-ahead buffer, while the search buffer is filled with zeros. The algorithm finds the longest match for the look-ahead buffer's prefix within the search buffer. The prefix \texttt{00} matches at multiple positions; an offset of \(p=0\) and a length of \(l=2\) are selected. The first mismatching character is \texttt{4}, yielding the output triple \((0, 2, 4)\). The buffers then advance by \(l+1=3\) characters.
    \item After the shift, the new content in the look-ahead buffer is \texttt{40040042}. The longest match found is \texttt{004004}, starting at offset \(p=5\) in the search buffer with length \(l=6\). The following character is \texttt{2}, producing the triple \((5, 6, 2)\). The buffers advance by 7 characters.
    \item The next character in the look-ahead buffer is \texttt{3}, which has no match in the current search buffer. This case is encoded as a literal with \(p=0\), \(l=0\), and \(c=3\), resulting in \((0, 0, 3)\). The buffers advance by 1 character.
    \item The final content in the look-ahead buffer is \texttt{04237}. The longest match is \texttt{0423}, found at offset \(p=4\) with length \(l=4\). The subsequent character is \texttt{7}, yielding the triple \((4, 4, 7)\). With the entire input processed, the algorithm terminates.
\end{enumerate}

The complete encoded output is the concatenation of the triples from each iteration: \texttt{024562003447}.


A cornerstone of LZ77 is its theoretical connection between the compression ratio and the information-theoretic entropy of the source. For a stationary information source \(\sigma\) over a finite alphabet \(A\), LZ77~\cite{ziv1977lz77} establishes the following bound when the buffer is sufficiently long:
\[
    h(L_s - 1) \le \rho \le h(L_s - 1) + \epsilon(L_s),
\]
where \(\rho\) is the achieved compression ratio, \(h(k)\) is defined as the per-symbol entropy \(h(k) = \log_{|A|}|\sigma\{k\}| / k\) with \(\sigma\{k\} = \{S \mid S \in \sigma \land |S| = k\}\), and \(\epsilon(L_s)\) is asymptotically \(\mathcal{O}(\log L_s / L_s)\). This classical result is fundamental to our analysis; therefore, we restate its proof in Appendix B for completeness. 

A simple rearrangement of this inequality yields
\[
    \rho - \epsilon(L_s) \le h(L_s - 1) \le \rho.
\]
This inequality confirms that the empirical compression ratio \(\rho\) provides a direct estimate---bounded by a vanishing term \(\epsilon(L_s)\)---of the true per-symbol entropy \(h(L_s-1)\) of the source. This result provides a rigorous, information-theoretic foundation for using compression ratios as quantitative proxies for information content in our analysis.

Applying the LZ77 algorithm, we compress all model outputs from our empirical study (Section~\ref{sec-study}). The average compression ratio computed across different stages of the generation process is presented in Figure~\ref{fig-theory-compression-eval}. These results allow us to quantify how information content evolves---or stagnates---during prolonged decoding under sparse attention, directly testing our initial observation.

\section{Theoretical Analysis for LZ77}

To establish the relationship between the compression ratio and the entropy rate, we employ the following insight: since a limited information source has only constrained substrings, if the buffer length is sufficiently large to accommodate a substantial number of such substrings, then we can use the constraint to bound the compression ratio.

Consider a source $\sigma$ defined over a finite alphabet $A$, constituted by a collection of strings with the property that certain specific substrings are prohibited. For each integer $k$, let $\sigma\{k\}$ denote the set of all length-$k$ strings belonging to $\sigma$. The associated per-symbol entropy is given by $h(k) = \frac{1}{k}\log|\sigma\{k\}|$. Note that it suffices to derive a bound for strings in $\sigma\{n - L_s\}$, because any such bound automatically extends to longer strings.

Take an arbitrary message $M \in \sigma\{n - L_s\}$. Suppose that the algorithm partitions $M$ into substrings $M = m_1 m_2 \dots m_N$, which are subsequently encoded into codewords $c_1, c_2, \dots, c_N$ of fixed length $\lceil\log_{|A|}L_s\rceil + \lceil\log_{|A|}(n - L_s)\rceil + 1 \triangleq L_c$. The compression ratio is therefore $\rho = \frac{L_c N}{n - L_s}$. Consequently, estimating $\rho$ reduces to bounding $N$.

To estimate $N$, we analyze the partition in finer detail. For each $p$, let $K_p$ be the number of substrings among $m_1, \dots, m_{N-1}$ having length $p$. Then
\[
N = 1 + \sum_{m=1}^{L_s} K_m.
\]
Thus the problem translates to bounding the counts $K_m$. Notice that if two substrings share the same length, they must be distinct; i.e., $|m_i| = |m_j|$ implies $m_i \neq m_j$. So set $l = L_s - 1$ and define $\lambda = \lceil \log |\sigma\{l\}| \rceil$. We bound $K_m$ in three regimes:

\begin{enumerate}
    \item For $1 \le m \le \lambda$, the trivial bound $K_m \le |A|^m \triangleq K'_m$ holds.
    \item For $\lambda < m \le l$, any substring of length $m$ can be extended (in at least one way) to a string of length $l$ in $\sigma\{l\}$. Hence $K_m \le |\sigma\{l\}| \triangleq K'_m$.
    \item For $m = l+1$, we utilize the total length constraint
          \[
          n - L_s = |m_N| + \sum_{m=1}^{l+1} m K_m,
          \]
          which yields
          \[
          K_{l+1} \le \frac{1}{l+1} \Bigl( n - L_s - \sum_{m=1}^l m K_m \Bigr).
          \]
          Substituting the upper bounds $K'_m$ for $K_m$ on the right-hand side produces an bound $K'_{l+1}$ for $K_{l+1}$. Observing that the other terms \(K_m\ (m\le l)\) are individually overestimated, the bound on \(K_{l+1}\) obtained from the fixed total length constraint might be an underestimate. However, because each $K_m$ contributes equally to $N$ but the coefficient of $K_{l+1}$ in the length sum is largest, the overall effect of substituting the overestimated \(K'_m\) into the bound for \(K_{l+1}\) still yields an overestimate for the total \(N\).
\end{enumerate}

Collecting these bounds, we obtain
\[
N \le K'_{l+1} + \sum_{m=1}^l K'_m \triangleq N'.
\]

Now select $n$ as follows:
\[
\begin{aligned}
n = & \sum_{m=1}^{\lambda} m |A|^m + \sum_{m=1}^{\lambda} m |\sigma\{m\}| \\
    & + (l+1) \Bigl( \sum_{m=1}^{\lambda} (l-m) |A|^m \\
    &+ \sum_{m=1}^{\lambda} (l-m) |\sigma\{m\}| + 1 \Bigr).
\end{aligned}
\]
With this choice, we achieve $N' = \frac{n - L_s}{l}$, leading to the compression ratio bound
\[
\rho \le \frac{L_c}{l} = \frac{L_c}{L_s - 1}.
\]
A trivial lower bound is $\rho \ge \frac{L_c}{L_s}$; hence, the derived upper bound is reasonably tight.

Furthermore, note that the codeword length satisfies $L_c \le 3 + \log(L_s - 1) + \log(n - L_s)$. From the definition of $n$,
\[
\begin{aligned}
n - L_s = & \; l \cdot \Bigg[ \sum_{m=1}^{\lambda} (l-m) |A|^m \\
&+ \sum_{m=\lambda+1}^{l} (l-m) |\sigma\{l\}| \\
          & + \sum_{m=1}^{\lambda} |A|^m + \sum_{m=\lambda+1}^{l} |\sigma\{l\}| \Bigg].
\end{aligned}
\]
Using the inequality $|A|^m \le |\sigma\{l\}|$ for all $m \le l$, we simplify to obtain
\[
n - L_s \le \frac{1}{2} \, l^2 (l+1) \, |\sigma\{l\}|.
\]
Substituting the preceding into the bound for $L_c$ gives
\[
L_c \le (L_s - 1) \Bigl[ h(L_s - 1) + \epsilon(L_s) \Bigr],
\]
where the error term is
\[
\epsilon(L_s) = \frac{1}{L_s - 1} \Bigl( 3 + 3\log(L_s - 1) + \log\frac{L_s}{2} \Bigr).
\]

\balance
\section{Checklist-Related Issues}

Three datasets, GSM8K (MIT), MATH500 (MIT), AIME (MIT), and three models, DeepScaleR 1.5B Preview (MIT), DeepSeek-R1-Distill-Llama-8B (MIT), and Qwen1.5-MoE-A2.7B-Chat (tongyi-qianwen) are used with their intended usage scenarios. We retrieve all models and datasets from Hugging Face, where detailed documentation, including parameter sizes and model architectures, is provided. We manually check the data and believe that there is no personal information misused.

We use ChatGPT to check the grammar of the texts.

To the best of our knowledge, we believe that our work does not pose risks that harm any subgroup of our society.

\end{document}